# Ranking Based Locality Sensitive Hashing Enabled Cancelable Biometrics: Index-of-Max Hashing

Zhe Jin, Yen-Lung Lai, Jung Yeon Hwang, Soohyung Kim, Andrew Beng Jin Teoh

*Abstract*—In this paper, we propose a ranking based locality sensitive hashing inspired two-factor cancelable biometrics, dubbed "Index-of-Max" (IoM) hashing for biometric template protection. With externally generated random parameters, IoM hashing transforms a real-valued biometric feature vector into discrete *index* (max ranked) hashed code. We demonstrate two realizations from IoM hashing notion, namely Gaussian Random Projection based and Uniformly Random Permutation based hashing schemes. The discrete indices representation nature of IoM hashed codes enjoy several merits. Firstly, IoM hashing empowers strong concealment to the biometric information. This contributes to the solid ground of non-invertibility guarantee. Secondly, IoM hashing is insensitive to the features magnitude, hence is more robust against biometric features variation. Thirdly, the magnitude-independence trait of IoM hashing makes the hash codes being scale-invariant, which is critical for matching and feature alignment. The experimental results demonstrate favorable accuracy performance on benchmark FVC2002 and FVC2004 fingerprint databases. The analyses justify its resilience to the existing and newly introduced security and privacy attacks as well as satisfy the revocability and unlinkability criteria of cancelable biometrics.

## I. INTRODUCTION

LATELY, rapid proliferation of biometric applications leads to massive amount of biometric templates. Public worries about the security and privacy of the biometric templates if stolen or compromised. Such concerns are attributed to the strong binding of individuals and privacy, and further complicated by the fact that biometric traits are irrevocable. Given the above threats, a number of proposals have been reported to protect the biometric templates. Generally, the proposals available in the literature can be broadly divided into two categories: *feature transformation* (or cancelable biometrics) and *biometric cryptosystems* (biometric encryption). Biometric cryptosystem serves the purpose of either securing the secret using biometric feature (key binding) or generating the secret directly from the biometric feature (key generation). On the other hand, cancelable biometrics [1] is truly meant for biometric template protection. It refers to the irreversible transform that can alter the biometric templates such that security and privacy of the templates can be assured. If a cancelable template is compromised, a new template can be re-generated from the same biometrics.

The cancelable biometric schemes in the literature vary according to different biometric modalities. However, the operation of a general cancelable biometrics system is similar to the conventional biometric system where the system composes of sensor, feature extractor and matcher except the former includes a parameterized transformation function right after feature extractor, and the matching is done in the transformed domain, rather in the feature domain. The parameters can be the passwords or user-specific seeded (pseudo) random numbers. The transformation function can be either a many-to-one function (non-invertible transforms) or a blending mechanism (salting) to alter the biometric templates [2]. A block diagram for general cancelable biometrics system is depicted in Fig. 1.

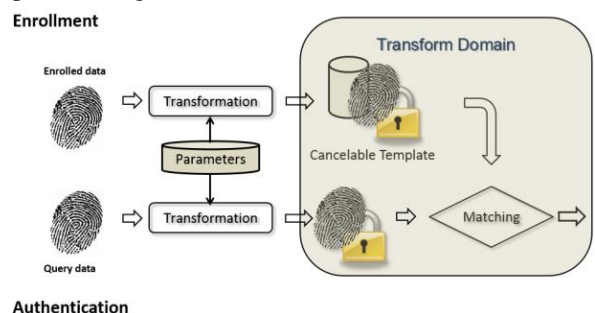

Fig. 1: A block diagram for general cancelable biometric systems.

Fingerprint is probably the most widely used biometric trait and the fingerprint minutiae is deemed as a reliable and stable local feature for accurate verification and identification [3]. It was believed that fingerprint image reconstruction from its minutiae is infeasible, but was disproved otherwise by Feng & Jain [4]. It is indeed not secure to store the original minutiae as

This work was supported by Institute for Information & communications Technology Promotion (IITP) grant funded by the Korea government (MSIT) (No.2016-0-00097, Development of Biometrics-based Key Infrastructure Technology for On-line Identification).

Zhe Jin and Yen-Lung Lai are with School of Information Technology, Advanced Engineering Platform (AEP), Monash University Malaysia ({Jin.Zhe, Lai.yenlung@monash.edu).

Jung Yeon Hwang and Sooyung Kim are with Electronics and Telecommunications Research Institute (ETRI) ({videmot, lifewsky}@etri.re.kr).

Andrew Beng Jin Teoh is with the School of Electrical and Electronic Engineering, College of Engineering, Yonsei University, Seoul, South Korea, (Corresponding author, e-mail: bjteoh@yonsei.ac.kr).

a template. Yet, it is quite challenging to design a template protection scheme with the following criteria [5].

- *Non-invertibility* or *Irreversibility*: It should be computationally infeasible to derive the original biometric template from a single instance or multiple instances of protected templates with or without auxiliary data (helper data). This property prevents the privacy attacks such as inversion attack and Attack via Record Multiplicity (ARM).
- *Revocability* or *Renewability*: A new instance of protected template can be revoked or re-issued when the old template is compromised.
- *Non-linkability* or *Unlinkability:* It should be computationally difficult to differentiate two or more instances of the protected templates derived from the same biometric trait. It prevents the cross-matching across different applications, thus, preserves the privacy of individual.
- *Performance preservation*. The accuracy performance of the protected templates should be preserved with respect to the before-transformed counterparts.

The rest of the paper is organized as follows. First, literature is reviewed in Section II. In Section III, the motivations and contributions of the paper are highlighted. In Section IV, the relevant background knowledge is given. The IoM hashing is described in Section V. Next, the experimental results are given in Section VI, supported by the performance analysis. In Section VII, security, privacy and revocability analysis are drawn. Finally, conclusion is given in Section VIII.

## II. Literature Review

In the past decades, numerous template protection schemes have been proposed. Several decent and comprehensive review papers exist in this topic such as [2] [5] [6]. We direct the readers to explore the details from these review papers.

In this paper, we focus on review 1) several state-of-the-arts fingerprint minutia-based cancelable template schemes since fingerprint modality is the subject of study in this paper; 2) a number of *generic* cancelable biometric schemes. By generic, we refer to the schemes that are applicable to other biometric modalities such as face and iris as IoM hashing is indeed a kind of generic cancelable biometric scheme.

### A. Minutiae-based Fingerprint Cancelable Template

Yang & Busch [7] proposed a template protection method based on the fingerprint minutia vicinity. Given $N$ minutiae $\{m_i \ /i=1,...,N\}$, each minutia $m_i$ with the three nearest neighboring minutiae $\{c_{i1},c_{i2},c_{i3}\}$ together form a set of minutia vicinity $V_i =\{(m_i,c_{i1},c_{i2},c_{i3})/i=1,...,N\}$. Each minutia vicinity comprises of 12 orientation vectors: $m_i \rightarrow c_{i1}$, $c_{i2} \rightarrow c_{i3}$, $c_{i3} \rightarrow c_{i1}$, etc. The four coordinate pairs of $V_i$ are then transformed based on the 5 (out of 12) randomly selected orientation vectors in the respective minutia vicinity. Next, the random offsets are added to each $V_i$ in order to conceal the local topological relationship among the minutiae in the vicinity. The transformed minutiae are thus regarded as a protected minutia vicinity with stored random offsets.

However, Simoens et al. [8] pointed out that the coordinates and orientations of minutiae in [7] could be revealed if both random offsets and orientation vectors are disclosed to an adversary. They also showed that the attack complexity is considerably low (e.g., only $2^{17}$ attempts are required when the random offsets table is known with reference to $2^{120}$ attempts when the random offsets table absent). Although Yang et al. [9] later proposed a dynamic random projection that was originally outlined in Teoh et al. [10], to alleviate this problem, dynamic random projection incurs substantially increased computation cost than that of random offsets used in [7].

Ferrara et al. [11] demonstrated a recovery algorithm to reveal the original minutiae from the minutia cylinder-code (MCC) descriptor, a state-of-the-art fingerprint descriptor proposed by Cappelli et al. [12]. A non-invertible scheme for MCC is hence proposed, namely protected minutia cylinder-code (P-MCC). The P-MCC computes the mean vector $\bar{x}$ and $k$ largest eigenvectors $\Phi_k$ from the MCC descriptor during training. These parameters are used for Binary-KL projection, which can be viewed as a one-bit binarization process of the projected features. Nevertheless, the cancellability is not addressed in [11]. A two-factor protection scheme based on P-MCC, namely 2P-MCC [13] is put forward to make the P-MCC be cancelable. Yet, the specific design of the MCC and its variations for point set data limit the propagation to other popular biometric modalities such as face and iris.

Multi-line Code (MLC) proposed by Wong et al. [14] is a minutia descriptor constructed based on the spatial distribution of the neighboring minutiae within a fixed radius. Firstly, a straight line is drawn following the direction of the reference minutiae and a number of overlapped circles with a pre-defined radius can be constructed. Then the neighbor minutiae are separated into different bins according to their orientation. Compute the mean of the distances between the center of the circle and the included minutiae for each region. In the binarization stage, two methods are used viz 1-bit and $k$-bits binarization. 1-bit binarzation is implemented based on a threshold value while gray code is used in $k$-bits implementation. If a MLC is compromised, the adversary can only obtain the rotated version of minutia vector. The regions reconstructed from different minutia vectors are uncorrelated as the orientations of the reference minutiae are independent to each other. Thus, the MLC may reduce the effort of brute force attack but difficult to recover the minutiae set.

Wang and Hu [15] proposed a blind system identification approach for biometric template protection. This is motivated by the fact that source signal cannot be recovered if the identifiability is dissatisfied in blind system identification. This approach exhibits decent accuracy performance preservation and the irreversibility of transformed template is justified theoretically and experimentally. However, the protected template against other major attacks (e.g. ARM) is unknown.

### B. Generic Cancelable Biometric Techniques

A well-known instance of salting based generic cancelable biometrics scheme, namely Biohashing [16], is based on the user-specific random projection where the distance preservation is proven by the J–L lemma [17]. Briefly, a user-specific orthonormal random matrix $R \in \mathbb{R}^{n \times q}$ where $q \leq n$ is generated from a seeded pseudo-random number generator. The extracted feature vector $\mathbf{x} \in \mathbb{R}^n$ is then projected via $\mathbf{y} = R^T \mathbf{x}$, and $y_i \in \mathbf{y}$ is binarized based on a chosen threshold value $\tau$ such as: $b_i = 0$, if $y_i < \tau$, and $b_i = 1$ otherwise, for $i = 1, \ldots, q$. In the event of template is compromised, a new template can be generated by issuing a new set of random vectors from the user-specific token. Biohashing works in various biometric traits such as fingerprint minutia [9], fingerprint texture [16], face [10], iris [18] etc. However, the non-invertibility of the Biohashing could be jeopardized if both $\mathbf{y}$ and $R$ are revealed. This is because Biohash is essentially a quantized under-determined linear equation system, which could be solved partially via pseudo-inverse operation [10].

Another work on cancelable iris that follows the random projection notion is Pillai et al. [19]. In this work, Gabor features are first computed from the iris pattern of the user. Random projections are then applied separately on each iris sector and the resulting transformed vectors are concatenated to form a cancelable template. Notably, [19] also reveals that a given random matrix $A \in \mathbb{R}^{n \times d}$ with $n \ll d$ drawn from a standard Gaussian leads to small restricted isometry constants $\delta_S$ with 'overwhelming probability' as far as $n = O(S \log(d/S))$ [20], given the iris features $\mathbf{x}$ has at most $S$ non-zero components. After projection with $A$, the probability to reconstruct $\mathbf{x}$ decreases if $S$ is large. More works that utilizing random projection can be found in [21] [22] [23].

Recently, another generic cancelable biometrics technique, namely Bloom filter has been introduced. The Bloom filter demonstrates a well adaptation to the popular biometric modalities such as iris [24], face [25] and fingerprint [26]. This technique is first reported for iris [24] as follows: the Bloom filter $b$ is a binary vector of length $n$. The Bloom filter is initialized with zeros and formed by adding elements '1' into it using $K$ independent hash function $\{h_i \in [0, n-1] | i = 1, \ldots K\}$. Practically, instead of using $K$ independent hash functions, a binary-to-decimal mapping is proposed. The IrisCode with dimension $H \times W$ is first splited into $K$ blocks with size $l = \frac{W}{K}$, where $l$ is the number of columns of each block. Each block $B_i$, $i \in [1, K]$ constitutes the formation of the bloom filter $b_i$. This can be done by adding element '1' to $b_i$ based on the position that manifested by each column codeword $x_j \in \{1,0\}^m$ inside $B_i$, where $j \in [1, l]$ and $1 \leq m \leq H$. Same $x_j$ will be mapped to the same element in the bloom filter resulted a many-to-one mapping hence non-invertibility criterion is satisfied. To achieve cancellability, an application-specific parameter $T$ is applied, just like other cancelable biometric schemes do.

However, despite decent performance preservation, security and privacy of the Bloom filter based schemes remains unsolved. For instance, Hermans et al. [27] demonstrates a simple and effective attack scheme that matches two protected templates derived from the same IrisCode using different secret bit vectors, thus break the requirement of non-linkability. Moreover, a security analysis also reveals that the false positive or key recovery can be accomplished with a low attack complexity of $2^{25}$ and $2^2$ to $2^8$ attempts respectively [27]. Bringer et al. [28] further analyzed the non-linkability of the protected templates generated from two different IrisCode of the same subject. They revealed when the key space is too small, a brute force attack would be succeeded while the accuracy performance declines if the key size is increased. Experiment confirms the vulnerability of irreversibility with block width of 16 or 32 in the Bloom filter scheme.

III. MOTIVATIONS AND CONTRIBUTIONS

From the literature, we can observe the limitations of present cancelable biometric schemes such as:

- Some of the afore-discussed "non-invertible transforms" are in fact susceptible to partial or full inversion (e.g., [1] [7] [16] [24]). For instance, many-to-one function applied in [1] degenerates to a linear function due to the improper parameters selection or when multiple protected templates are known to the adversary [15] [16].
- Some of the afore-discussed methods are weak to survive from the security and privacy attacks, e.g. ARM [15], false-accept attack (dictionary attack) [11].
- State-of-the-art fingerprint minutiae protection schemes (e.g. P-MCC [12], 2P-MCC [13]) limits the coverage to certain biometric feature/modality, i.e. point set data.
- Trade-off between performance and non-invertibility. This is due to the contradiction while strict non-invertibility implies total loss of information is required after transformation, accuracy performance preservation is only possible when discriminative information of the original template is conserved [5].
- Accuracy performance discrepancy. Some of the salting based cancelable biometric approaches that utilize user-specific parameter such as Biohashing [16] suffer from the accuracy performance discrepancy under genuine-token and stolen-token scenarios [1]. It is often observed that genuine-token scenario yields a huge performance gain over stolen-token scenario, which imposes an unrealistic assumption where the token is to be secretly kept all the time. Therefore, any template protection scheme that requires independent parameters to achieve revocability/cancellability should be free from this issue [5].

With above justifications, we outline a generic salting based cancelable biometric scheme dubbed Index-of-Max (IoM) hashing. The IoM hashing is motivated from the Locality Sensitive Hashing (LSH) in information retrieval domain. As a special instance of LSH family, the IoM hashing enjoys the merits of strong theoretical and empirical guarantee of accuracy preservation after hashing. With its pure discrete indices (max ranked) representation nature that non-linearly transformed from the real-valued biometric features, the IoM hashing can strongly protect the biometric data from being inverted. In this paper, we demonstrate the application of IoM hashing in fixed length fingerprint vector derived from fingerprint minutiae [29].

The main contributions of this paper are as follows:
1) We devise a novel ranking based LSH inspired IoM hashing as a means of two-factor cancelable biometric construct. With externally generated random parameters, the IoM hashing strongly conceals the biometric features while satisfies the accuracy performance, revocability and renewability criteria.
2) We put forward two realizations of the IoM hashing, namely, i.e. Gaussian Random Permutation (GRP)-based and Uniform Random Projection (URP)-based hashing schemes and justify their characteristics both qualitatively and quantitatively. The two realizations are also shown free from accuracy performance discrepancy problem.
3) We rigorously analyze the security and privacy aspects of the IoM hashing in both qualitative and quantitative manners. The specific instances include non-invertibility and non-linkability analysis. Moreover, the existing major privacy and security attacks e.g. invertibility attack, brute force attack, false accept attack, ARM attacks against IoM hashing are highlighted.
4) We introduce a new sophisticated attack for cancelable biometric schemes, namely birthday attack, which is more vulnerable than the false accept attack.

## IV. PRELIMINARIES

In this section, we give a brief account for Locality Sensitive Hashing (LSH), in which our IoM hashing based upon. Besides that, we also present the "Winner Takes All" (WTA) hashing [30] and the "Random MaxOut Features" (RMF) [31] that have inspired us in devising URP-based and GRP-based IoM hashing, respectively. WTA hashing was meant for data retrieval while RMF is for data classification. The latter is indeed a spinoff from maxout network in machine learning.

### A. Locality Sensitive Hashing

Locality sensitive hashing (LSH) is primarily used to reduce the dimensionality of high-dimensional data by hashing the input items so that similar items map to the same "buckets" with high probability where the number of buckets being much smaller than the input items. The chief objective of LSH is to maximize the probability of a "collision" for similar items. Formally, the LSH family $H$ is defined as follows:

**Definition 1** [32]: *A LSH is a probability distribution on a family H of hash functions h such that* $\mathbb{P}_{h \in H}[h(X) = h(Y)] = S(X, Y)$ *where S is a similarity function defined on the collection of object X and Y.*

The key ingredient of LSH is the hashing of object collection $X$ and $Y$ by means of multiple hash functions $h_i$. The use of $h_i$ enables decent approximation of the pair-wise distance of $X$ and $Y$ in terms of collision probability. LSH ensures that $X$ and $Y$ with high similarity renders higher probability of collision in the hashed domain; on the contrary, the data points far apart each other result a lower probability of hash collision.

$$\begin{aligned} \mathbb{P}_{h \in H}\big(h_i(X) = h_i(Y)\big) \leq P_1, & \text{ if } S(X,Y) < R_1 \\ \mathbb{P}_{h \in H}\big(h_i(X) = h_i(Y)\big) \geq P_2, & \text{ if } S(X,Y) > R_2 \end{aligned} \quad (1)$$

Given that $P_2 > P_1$, while $X, Y \in \mathbb{R}^d$, and $H = \{h: \mathbb{R}^d \to U\}$, where $U$ is a hashed metric space depending on the similarity function defined by $S$, $i$ refers to the number of hash functions $h$.

### B. "Winner-Takes-All" Hashing

The basic idea of WTA hashing [30] is to compute the ordinal embedding of an input data based on the partial order statistics. More specifically, the WTA is a non-linear transformation based on the *implicit order* rather than the absolute/numeric values of the input data, and therefore, offers a certain degree of resilience to numerical perturbation while giving a good indication of inherent similarity between the compared items [30]. The overall WTA hashing procedure can be summarized into five steps:
1. Perform $H$ random permutations on the input vector with dimension $d$, $\mathbf{x} \in \mathbb{R}^d$.
2. Select the first $k$ items of the permuted $\mathbf{x}$.
3. Choose the largest element within the $k$ items.
4. Record the corresponding index values in bits.
5. Step 1 – step 4 is repeated $m$ times, yielding in a hash code of length $m$, which can be compactly represented using $m\lceil log_2 k \rceil$ bits.

### C. "Random MaxOut Features"

RMF proposed by Mroueh et al. [31] is a simple yet effective non-linear feature mapping that approximates the functions of interest. Let $\{\mathbf{w}_j^i \in \mathbb{R}^d | i = 1, \dots, m, j = 1, \dots, q\}$ be the iid standard Gaussian random vector drawn from $\mathcal{N}(0, I_d)$. For $\mathbf{x} \in \mathbb{R}^d$, the RMF is defined as $\varphi_i(\mathbf{x}) = \max_{j=1\dots q} < \mathbf{w}_j^i, \mathbf{x} >, i = 1, \dots, m$. A collection of $m$ RMF yields a RMF vector $\Phi(\mathbf{x}) = \frac{1}{\sqrt{m}} [\varphi_1(\mathbf{x}), \dots, \varphi_m(\mathbf{x})] \in \mathbb{R}^m$.

## V. METHODOLOGY

The IoM hashing takes a fixed-length fingerprint vector (hereafter referred to as *fingerprint vector*) as input; thus we first give a brief description on fingerprint vector and then followed by the IoM hashing and its two realizations, i.e. Gaussian Random Projection (GRP) and Uniformly Random Permutation (URP). For quick reference, the notations are given in TABLE I:

### A. Globally ordered fixed-length fingerprint vector

TABLE I
NOMENCLATURE

| Notation(s) | Description |
|---|---|
| $H$ | Locality-sensitive hashing (LSH) family |
| $h$ | LSH function $h \in H$ |
| $S(X,Y)$ | Similarity function defined on the collection of object $X$ and $Y$ |
| $\mathbf{x}$ | Fingerprint vector $\mathbf{x} \in \mathbb{R}^d$ |
| $m$ | Number of Gaussian random matrices (GRP-based IoM)/ Number of random permutation (URP-based IoM) |
| $q$ | Number of Gaussian random projection vector |
| $p$ | Order of Hadamard product |
| $k$ | Window size (URP-based IoM hashing) |
| $t$ | $t_{GRP}$ - GRP-based hashed code, $t_{URP}$ - URP-based hashed code |

As outline in [29], the construction of fingerprint vectors, consists of three main steps, i.e. minutiae descriptor extraction, kernel learning-based transformation and feature binarization. Since IoM hashing takes a real-valued vector as input, binarization is hence dropped. The procedure of generating fingerprint vectors is given as follows (MATLAB codes available at goo.gl/8EoLsp):

1. Minutiae descriptor extraction, $\boldsymbol{\Omega}$: Minutia Cylinder-Code (MCC) [12] is used at this step. The MCC is meant to capture spatial and directional relation between the reference minutia $m_r = \{x_r, y_r, \theta_r\}$ and the neighbor minutiae $m^b = \{x_i^b, y_i^b, \theta_i^b | i = 1, \dots n_r - 1\}$, where $n_r$ is the total number of minutiae within a fixed-radius $r$.

2. Kernel matrix computation. Let $\boldsymbol{\Omega} = \{\boldsymbol{\Omega}^t(i) | i = 1, \dots, N_t\}$ be a set of MCC training samples and $N_t$ denotes the total number of $\boldsymbol{\Omega}^t$. A kernel matrix $\mathbf{K} \in \mathbb{R}^{N_t \times N_t}$ is then computed with the kernel function given in eq. (2), where $S_{MCC} \in [0,1]$ is the MCC dissimilarity measure and $\sigma$ is the spread factor.

$$\mathbf{K}(i,j) = k(\boldsymbol{\Omega}^t(i), \boldsymbol{\Omega}^t(j)) = \exp(-(1 - S_{MCC}(i,j))^2 / 2\sigma^2) \quad (2)$$

3. With $\mathbf{K}$, the projection matrix $\mathbf{P} \in \mathbb{R}^{N_t \times d}$, i.e. the eigenvectors of the kernel principal component analysis (KPCA) [33] can be inferred, where $d$ denotes the number of desired output dimensions.

4. Let $\boldsymbol{\Omega}^q$ be the MCC descriptor query instance. The $\boldsymbol{\Omega}^q$ is first matched with all training samples $\boldsymbol{\Omega}^t(i)$ for $1 \le i \le N_t$. Subsequently, a vector $\mathbf{v} \in \mathbb{R}^{N_t}$ can be formed by concatenating the $N_t$ matching scores $v_i$:

$$v_i \leftarrow sim(\boldsymbol{\Omega}^q, \boldsymbol{\Omega}^t(i)) \quad (3)$$

5. $\mathbf{v}$ is then transformed with the kernel function in eq. (2) yields

$$\bar{\mathbf{v}} = \exp(-(1 - \mathbf{v})^2 / 2\sigma^2) \quad (4)$$

6. A fingerprint vector $\mathbf{x}$ can be generated through:

$$\mathbf{x} = \bar{\mathbf{v}} \mathbf{P} \in \mathbb{R}^d \quad (5)$$

where $\bar{\mathbf{v}} \in \mathbb{R}^{N_t}$ and $\mathbf{P} \in \mathbb{R}^{N_t \times d}$.

B. *The IoM Hashing*

The IoM hashing is a means of cancelable biometrics that can be perceived as a special instance of LSH portray in Definition 1. The locality sensitive function $h(.)$ in our context, refers to the $q$-dimension random projection in GRP-based realization (section V(C)) and the $p$-order Hadamard product in URP-based IoM hashing (section V(D)), respectively. In general, the IoM hashing embeds the biometric features non-linearly onto a rank metric space that characterized by the LSH admitted ranking based similarity functions $S$. In this work, two different ranking based functions are identified for GRP and URP realizations (section V(E)). (MATLAB codes available at goo.gl/8EoLsp).

As a LSH instance, the accuracy performance of biometric features can be preserved largely after IoM hashing. The pure discrete indices representation nature of the IoM hashed code enjoys several merits as follows:

1. The IoM hashing empowers strong concealment to biometric information, which is always manifested in terms of feature magnitudes. This contributes to the solid ground of non-invertibility guarantee (refer section VII(A)).

2. The IoM hashing is essentially a type of ranking based hashing method that relies on the *relative ordering* of feature dimensions, and thus independent to the magnitude of the features. This makes the hashed codes robust against noises and variations that would not affect the implicit ordering. In fact, the works based on the ranking notion are not uncommon in computer vision [34] and information retrieval domain [30] [35].

3. Magnitude-independence of the IoM hashing makes the resultant hash codes being scale-invariant, which is critical to compare and align the features from heterogeneous spaces [35].

C. *Gaussian Random Projection (GRP)-based IoM hashing*

The GRP-based IoM hashing can be condensed into a two-step procedure as follows:

1. Given a fingerprint vector $\mathbf{x} \in \mathbb{R}^d$, generate $q$ number of random Gaussian projection vectors for $m$ times, $\{\mathbf{w}_j^i \in \mathbb{R}^d | i = 1 \dots, m, \ j = 1, \dots, q\} \sim \mathcal{N}(0, I_d)$ and hence a random Gaussian projection matrix $\mathbf{W}^i = [\mathbf{w}_1^i, \dots \mathbf{w}_q^i]$ can be formed.

2. Record the $m$ indices of the maximum value computed from $\varphi_i(\mathbf{x}) = \arg \max_{j=1 \dots q} \langle \mathbf{w}_j^i, \mathbf{x} \rangle$ as $t_i$. The GRP-based IoM hashed code is hence $\boldsymbol{t}_{GRP} = \{t_i \in [1, q] | i = 1, \dots, m\}$.

Pseudo-code of the GRP-based IoM hashing is also given in Algorithm 1.

| Algorithm 1. Gaussian Random Projection based IoM hashing |
|---|
| **Input** Feature vector $\mathbf{x}$ with $d$ dimensions, number of Gaussian random matrices $m$ and the number of Gaussian random projection vector $q$ |
| **Step 1**: Generate $m$ Gaussian random matrices $\mathbf{W}^i = (\mathbf{w}_1^i, \dots, \mathbf{w}_q^i)$ $i = 1, \dots, m$. <br> **Step 2**: Initialize $i^{th}$ hashed code $t_i = 0$. <br> **Step 3**: Perform *random projection* and record the *maximum index* in the projected feature vector. <br> For $k = 1: m$ <br>      $\bar{\mathbf{x}}^k = \mathbf{W}^k \mathbf{x}$ <br>      Find $\mathbf{x}_j^k = \max(\bar{\mathbf{x}}^k), j = 1, \dots, q$ <br>      Then $t_i = j$ ($j$ refers the index of $\bar{\mathbf{x}}^k$) <br> End for |
| **Output** Hashed code $\boldsymbol{t}_{GRP} = \{t_i | i = 1, \dots, m\}$ and $\boldsymbol{t}_{GRP} \in [1, q]$. |

Note that GPR-based IoM hashing differs from the RMF in section IV(C) in such a way that the representation form of the former encodes the indices of the maximum value of $\varphi_i(\mathbf{x})$, whereas the latter is mere the maximum numeric values of $\varphi_i(\mathbf{x})$. In a nutshell, the GRP embeds the fingerprint vector onto a $q$-dimension Gaussian random subspace and the index of max-valued projected feature is taken, which is equivalent to a hashed entry, $h(\cdot)$ in the rank space from LSH viewpoint. This process is repeated with $m$ independent Gaussian random matrices and yield a collection of $m$ IoM indices. Therefore, the GRP hashed codes enjoy the Euclidean pairwise distance preservation in the projected subspace $\mathbb{R}^q$. The LSH admitted distance/similarity function $S$ is determined by the Lemma 1 in section V(E).

Despite GRP-based IoM hashing and Biohashing [16] share a common ground where both adopt random projections, the distinctions of them are more apparent:

1. The theory of GRP-based IoM hashing is based on the LSH that embeds any finite subset in Euclidian space to $O(\epsilon^{-2}r)$ dimensions with $1 + \epsilon$ distortion within the neighborhoods with at most $r$ points [36]. While the Biohashing is stemmed from the J–L lemma [17], which embeds any *n*-points in Euclidian space to $O(\epsilon^{-2}\log n)$ dimensions with $1 + \epsilon$ distortion. In other words, the IoM hashing can be perceived as a *local manifestation* of J-L lemma [17], which works well for local distance preservation but not globally preserved as in the Biohashing scheme. Besides, accuracy performance of the Biohashing could be deteriorated in stolen-token scenario due to simplistic thresholding operation that endures information lost. This issue does not exist in the GRP-based IoM hashing as no thresholding is used.
2. *Multiple* (*m* to be exact) Gaussian projection matrices $\mathbf{W}^i$ (*m* independent hashing from LSH perspective) are used in the GRP-based IoM hashing whereas mere one *single* random orthogonal projection matrix is used in the Biohashing;
3. The GRP-based IoM hashed code is a collection of independent discrete *index* values while Biohashing is a *real-valued* feature vector;
4. The projection matrix in the Biohashing should be kept secret for maximal security and privacy protection; on the contrary projection matrices of the GRP-based IoM hashing is not necessary to be private, in which we will further justify in section VII.

*D. Uniformly Random Permutation (URP)-based IoM hashing*

The key ingredient of URP-based IoM hashing is the uniform random permutation that inherited from the WTA hashing and further strengthened by the Hadamard product operation. The overall transformation is illustrated in Fig. 2. Suppose a feature vector $\mathbf{x} \in \mathbb{R}^d$ and $m$ independent hash functions $\{h_i(\mathbf{x}) \in [1, k] | i = 1, \dots, m\}$ where each hash function consists of $p$-order Hadamard product. The URP-based IoM hashed code is thus generated by concatenating the outputs from $m$ independent hash functions. The 5-steps procedure of the URP-based IoM hashing is as follows and its pseudo code is presented in Algorithm 2.

1. **Random Permutation**: For each $h(\mathbf{x})$, generate a permutation set $\theta$, which is generated from $p$ independent random uniform permutation seeds. Then, $\mathbf{x}$ is permuted yield $p$ permuted vectors, $\{\hat{\mathbf{x}}_j \in \mathbb{R}^d | j = 1, \dots, p\}$.
2. **Hadamard Product Vector Generation**: Generate a $p$-order Hadamard product vector by performing the element-wise multiplication on $p$ permuted vectors i.e. $\bar{\mathbf{x}} = \prod_{j=1}^{p} (\hat{\mathbf{x}}_j) \in \mathbb{R}^d$.
3. For each $\bar{\mathbf{x}}$, select the first $k$ elements where $1 < k < d$.
4. Record the index of the maximum value in the first $k$ elements (or $k$-size window for brevity) and the index value is stored as $t \in [1, k]$.
5. Repeat step 1 to 4 with different permutations set $\{\theta_{(l,i)} | l = 1, \dots, q, i = 1, \dots, m\}$ and construct URP based

**Algorithm 2. Uniformly Random Permutation based IoM hashing**

**Input** Feature vector $\mathbf{x}$, $p$ order of Hadamard product, number of random permutation $m$, window size $k$
For each permutation set $\{\theta_{(l,i)} | l = 1, \dots, q, i = 1, \dots, m\}$
**Step 1**: Permute elements in $\mathbf{x}$ based on $\theta_{(i,l)}$, $\hat{\mathbf{x}} = perm(\mathbf{x})$, $perm(.)$ is the random permutation function
**Step 2**: Initialize $i^{th}$ hashed code $t_i = 0$
**Step 3**: Hadamard product vector generation and output hashed codes
For $j$=1: $k$
    Set $\bar{\mathbf{x}}(j) = \prod_{l=1}^{p} (\hat{\mathbf{x}}_l(j))$
    if $\bar{\mathbf{x}}(j) > \bar{\mathbf{x}}(t_i)$ then $t_i = j$
End for
**Output** Hashed code $t_{URP} = \{t_i \in [1, k] | i = 1, \dots, m\}$

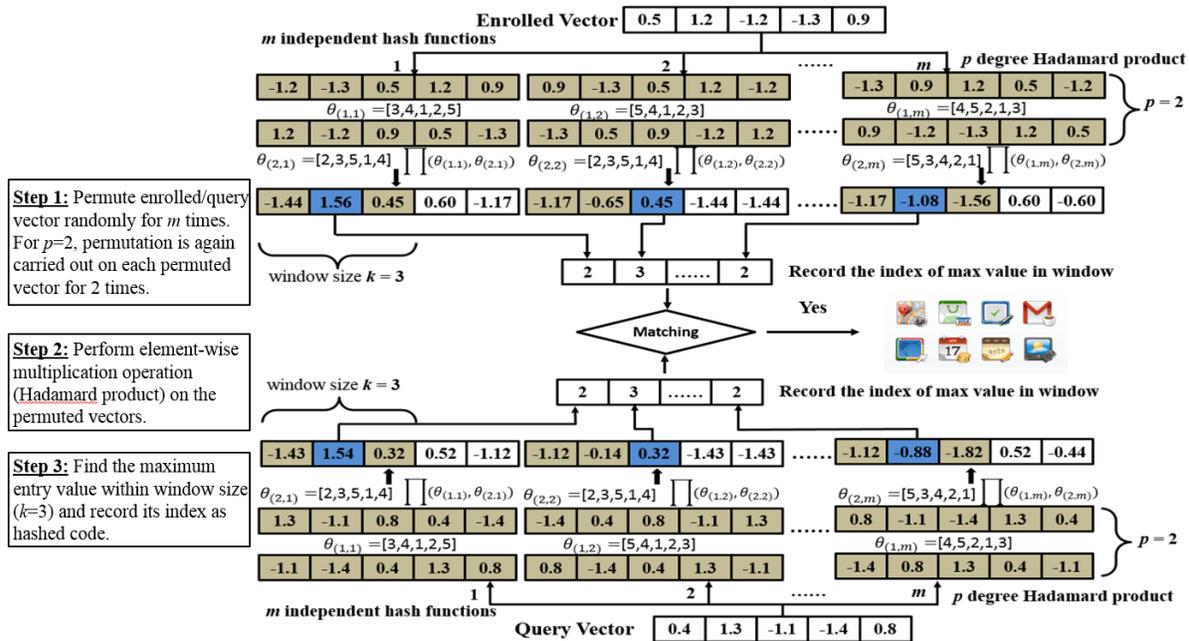

Fig. 2. Uniformly Random Permutation (URP) based IoM hashing where $d$=5, $p$=2 and $k$=3 are used as illustration.

IoM hashed code, $\boldsymbol{t}_{URP} = \{t_i \in [1,k] | i = 1, \ldots, m\}$.

Note that window size $k$ has an impact in the distribution of which indices from the fingerprint vector end up in the code. The choice of $k$ indeed leads to different emphasis on pair-wise agreements for indices at the first $k$ entries in the permuted vector. The value of $k$ offers a way to tune for giving higher weight to top entry in the biometric vectors versus the others [30].

We highlight that $p$-order Hadamard product in the URP-based realization is a crucial factor for security and privacy. Due to the $k$-windowed permutation is finite in principle, the adversary may launch brute force attack, dictionary attack or ARM (section VII(A) and VII(B)) to either gain illegal access to the system or to recover the biometric features. If both IoM hashed codes and permutation seeds are revealed, the attack complexity could be significantly reduced. The $p$-order Hadamard product can largely increases the difficulty of such recovery. We refer the readers to section VII(B)(2) for detail analysis.

*E. Matching of IoM Hashed Codes*

The IoM hashing essentially follows the ranking based LSH that strives to ensure that two fingerprint vectors with high similarity renders higher probability of collision in the rank domain; on the contrary, the vectors far apart each other result a lower probability of hash collision. Suppose two hashed codes, enrolled $\boldsymbol{t}^e = \{t_i^e | i = 1, \ldots, m\}$ and query $\boldsymbol{t}^q = \{t_j^q | j = 1, \ldots, m\}$ and $S(\boldsymbol{t}^e, \boldsymbol{t}^q)$ represents the *probability of collision* of two hashed codes i.e. $\mathbb{P}[t_i^e = t_j^q] = S(\boldsymbol{t}^e, \boldsymbol{t}^q)$ for $i = 1, \ldots, m$. The high probability of collision implies high similarity of $\boldsymbol{t}^e, \boldsymbol{t}^q$ and vice-versa.

It is important to note that despite both URP and GRP based IoM hashing encode the maximum values of the transformed features in indices representation, the respective rank similarity functions $S(\cdot,\cdot)$ are fundamentally different. In what follows, we first give a lemma that enables us to establish the GRP-based IoM matcher.

**Lemma 1** [37]: For $\boldsymbol{u}, \boldsymbol{v} \in \mathbb{R}^d$ be the unit vector $\|\boldsymbol{u}\| = \|\boldsymbol{v}\| = 1$ at angle $\theta$. Let $\rho = \boldsymbol{u} \cdot \boldsymbol{v} = \cos\theta$ and $\boldsymbol{r}_1, \ldots, \boldsymbol{r}_q$ be a sequences of iid standard Gaussian random vectors, the probability for $\boldsymbol{u}$ and $\boldsymbol{v}$ be not separated by $\boldsymbol{r}_1, \ldots, \boldsymbol{r}_q$ is designated as $k_q(\boldsymbol{u}, \boldsymbol{v})$. The Taylor series of $k_q(\boldsymbol{u}, \boldsymbol{v})$ is given:

$$k_q(\boldsymbol{u}, \boldsymbol{v}) = \sum_{i=0}^{\infty} a_i(q)\rho^i \quad (6)$$

where $k_q$ around $\rho = 0$, converges for all $\rho$ in the range of $|\rho| \leq 1$. The coefficients $a_i(q)$ are non-zero and their sum converges to 1. The first three coefficients can be expressed as $a_0(q) = \frac{1}{q}$, $a_1(q) = \frac{h_1^2(q)}{q-1}$, and $a_2(q) = \frac{qh_2^2(q)}{(q-1)(q-2)}$, where $h_i(q)$ is the expectation of $\phi_i(x_{max})$ where $\phi_i(\cdot)$ be the normalized Hermite polynomials and $x_{max}$ is the maximum entry of $q$ iid standard Gaussian random variables.

**Remark 1** [31]: Let $h: \mathbb{R}^d \to \{1, \ldots, q\}^m$ a hashing function that admits LSH, for $\boldsymbol{u}$ and $\boldsymbol{v} \in \mathbb{R}^d$ and $h(\boldsymbol{u}) = \{\arg\max_{j=1\ldots q}\langle \boldsymbol{w}_j^i, \boldsymbol{u}\rangle | i = 1, \ldots, m\}$ where $\{\boldsymbol{w}_j \in \mathbb{R}^d | j = 1, \ldots, q\} \sim \mathcal{N}(0, \mathbf{I}_d)$. The expectation of the probability that the identical entries appear in two hashed codes, can be estimated as $\mathbb{E}(\mathbb{P}[h_i(\boldsymbol{u}) = h_i(\boldsymbol{v})]) = \mathbb{P}\{\arg\max_{j=1\ldots q}\langle \boldsymbol{w}_j^l, \boldsymbol{u}\rangle = \arg\max_{j=1\ldots q}\langle \boldsymbol{w}_j^l, \boldsymbol{v}\rangle\}$. As $m$ becomes large, $k_q(\boldsymbol{u}, \boldsymbol{v})$ can be approximated by:

$$\mathbb{P}\left\{\arg\max_{j=1\ldots q}\langle \boldsymbol{w}_j^i, \boldsymbol{u}\rangle = \arg\max_{j=1\ldots q}\langle \boldsymbol{w}_j^i, \boldsymbol{v}\rangle\right\} \approx k_q(\boldsymbol{u}, \boldsymbol{v}) \quad (7)$$

$$S_{GRP}(\boldsymbol{t}^e, \boldsymbol{t}^q) = \mathbb{P}\left\{\arg\max_{j=1\ldots q}\langle \boldsymbol{w}_j^i, \boldsymbol{u}\rangle = \arg\max_{j=1\ldots q}\langle \boldsymbol{w}_j^i, \boldsymbol{v}\rangle\right\} \approx k_q(\boldsymbol{u}, \boldsymbol{v}) \quad (8)$$

The matching of GRP-based realization hence can be carried out based on (8), which echoes the probability of collision of two hashed codes. Operationally, the matching score is mere a total number of collisions which can be sought by counting the number of '0' (collisions) after the element-wise subtraction of $\boldsymbol{t}^e, \boldsymbol{t}^q$ over $m$ (total entries of the hashed code).

For URP-based IoM, the $S_{URP}(\cdot,\cdot)$ is a rank correlation measurement. Briefly, the rank correlation refers to the measurement of ordinal association ordinal measure) that based on the relative ordering of values in a given range. The pairwise-order $PO$, the simplest similarity measure for rank correlation is defined as [38]:

$$PO(\boldsymbol{t}^e, \boldsymbol{t}^q) = \sum_i \sum_{j<i} Q\left((t_i^e - t_j^e)(t_i^q - t_j^q)\right) = \sum_i R_i(\boldsymbol{t}^e, \boldsymbol{t}^q)$$

where $R_i(\boldsymbol{t}^e, \boldsymbol{t}^q) = |L(\boldsymbol{t}^e, i) \cap L(\boldsymbol{t}^q, i)|$
$L(\boldsymbol{t}^e, i) = \{j | \boldsymbol{t}^e(i) > \boldsymbol{t}^e(j)\}$
$L(\boldsymbol{t}^q, i) = \{j | \boldsymbol{t}^q(i) > \boldsymbol{t}^q(j)\}$
$Q(z) = \begin{cases} 1 & z > 0 \\ 0 & z \leq 0 \end{cases}$

The operational $PO$ function can be reformulated as in eq. (9) [30]:

$$S_{URP}(\boldsymbol{t}^e, \boldsymbol{t}^q) = PO(\boldsymbol{t}^e, \boldsymbol{t}^q) \approx \frac{\sum_{i=0}^{n-1}\binom{R_i(\boldsymbol{t}^e, \boldsymbol{t}^q)}{k-1}}{\binom{d}{k}} \quad (9)$$

An interpretation of eq. (9) is as follows: the enumeration in which index $i$ can be the max over a $k$-sized permuted window is given by the number of ways in which one can pick $k$-1 entries that are smaller than the entry at $i$ and common to both $\boldsymbol{t}^e$ and $\boldsymbol{t}^q$, in which also indicate the collision probability of a pair of URP-based hashed codes [30]. This also validates the LSH property of URP realization. Similarly, the matching score is mere by counting the number of 0 after the element-wise subtraction of $\boldsymbol{t}^e$ and $\boldsymbol{t}^q$ over $m$.

*F. Cancelable Template Generation*

In the event of template compromised, a IoM hashed code can be easily replaced with a new tokenized random permutation seed or tokenized Gaussian random matrices for URP or GRP, respectively. The effectiveness of the revocability is experimentally verified in section VII(D). In real world scenario, the random permutation seeds or random matrices are user-specific for revocability. However, stolen token/seed scenario should be focused as it is closely associated to accuracy performance, security and privacy attacks [5]. To evaluate the stolen-token scenario, our experiment is performed with same random token (i.e. random permutation seed or random matrices) for all subjects. The accuracy performance in stolen-token scenario is presented in VI(*B*).

### G. Generic Template Protection

The IoM hashing is generic in the sense that it can apply to most of the common biometrics in which their features are in binary form (e.g. iris, palmprint) and fixed-length vector form (e.g. face). Real-valued feature is directly applicable to the IoM hashing as demonstrated in this paper. A recent binary cancelable iris exposition, namely Index First One hashing (IFO) [39] is indeed a special case of the URP-based IoM hashing where the position of the first one in IFO is encoded in place of the position of maximum value in the URP-based IoM hashing. Attribute to the internal connection between IoM hashing and LSH, it is anticipated that the IoM hashing can be extended to other common biometric modalities.

## VI. EXPERIMENTS AND DISCUSSIONS

In this paper, we adopt fingerprint vector **x** with length 299 outline in [29] as an input for IoM hashing evaluations. The evaluations are conducted on six public fingerprint datasets, FVC2002 (DB1, DB2, DB3) [40] and FVC2004 (DB1, DB2, DB3) [40]. Each dataset consists of 100 users with 8 samples per user. There are 800 (100×8) fingerprint images in total in each dataset. The performance accuracy of the proposed method is assessed with Equal Error Rate (EER) and the genuine-imposter distributions. Note that since the random permutation/projection is applied, the EERs are calculated by taking the average of EERs repeated for five times.

For matching protocol, as described in [29], 1st to 3rd samples of each identity are used as training samples to generate the fingerprint vector; the rest of the samples (i.e. 4th – 8th) of each identity are used in this experiment. There are 500 (100×5) in total. Within this subset of data, the Fingerprint Verification Competition (FVC) [40] protocol is applied across the six data sets, which yields 1000 genuine matching scores and 4950 imposter matching scores for each data set.

### A. Parameters of IoM hashing

*1) Effect of window size k, Hadamard product order p, and number of hashing functions m*

For URP-based IoM, we first investigate the effect of window size $k$ with respect to the performance in terms of EER. In this experiment, $k$ is varied from 50, 80, 100, 128, 156, 200 and 250 by fixing $m$ at 600. The identical setting is repeated for $p = [2,3,4,5]$. Fig. 3 shows the curves of "EER (%)-vs-$k$" on FVC2002 (DB1, DB2) and FVC2004 (DB3). We can observe that:

1) The EER drops gradually when larger $k$ is applied and levels off when $k$ becomes big. This is not a surprise as small $k$ implies fewer comparisons and thus less informative of max ranked features.

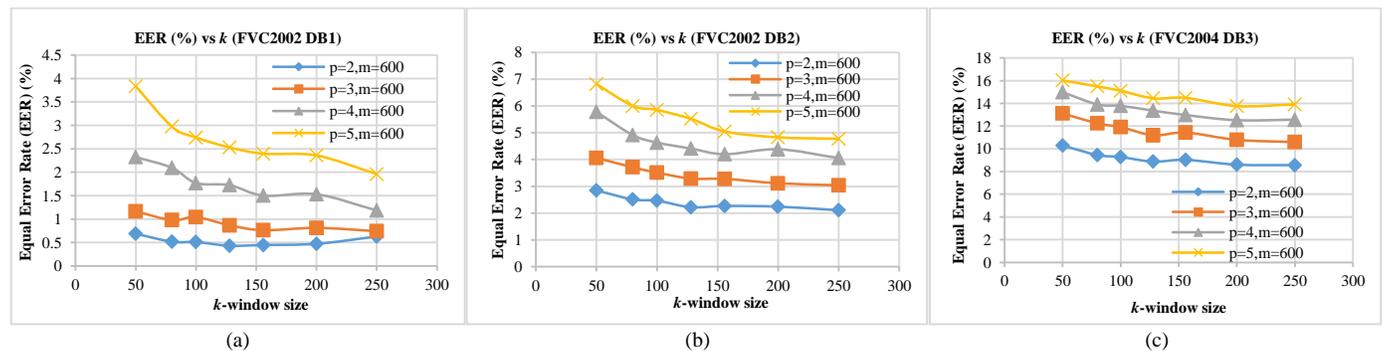

Fig. 3 The curves of "EER (%) vs $k$" on FVC2002 and FVC2004 (DB1, DB2, DB3)

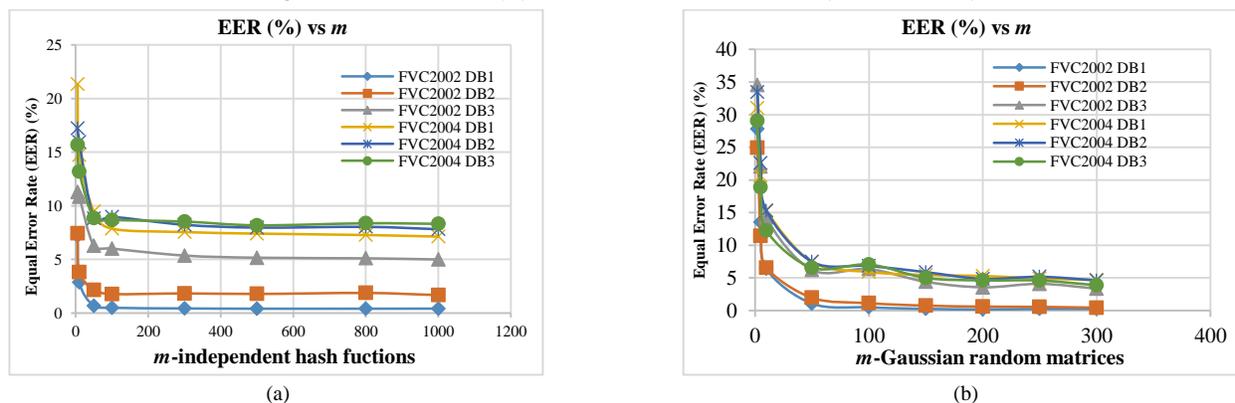

Fig. 4 (a) The curves of "EER (%) vs $m$-independent hash functions" FVC2002 and FVC2004 (DB1, DB2, DB3);
(b) The curves of "EER (%) vs $m$-Gaussian random matrices" FVC2002 and FVC2004 (DB1, DB2, DB3).

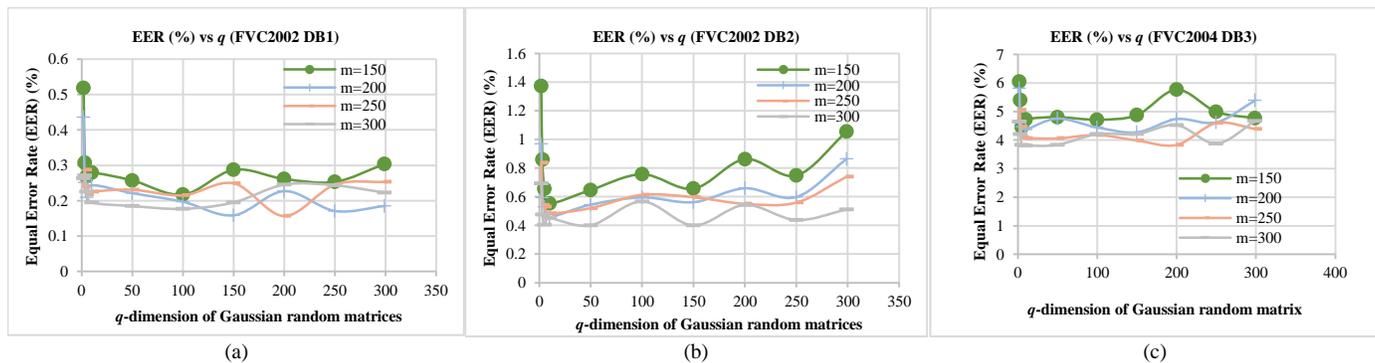

Fig. 5 The curves of "EER (%) vs $q$" on FVC2002 (DB1, DB2) and FVC2004 (DB3)

2) The smaller $p$, the lower EER. As indicated in Algorithm 2 (step 3), the Hadamard product, $\bar{\mathbf{x}}(j) = \prod_{l=1}^{p} (\hat{\mathbf{x}}_l(j))$ is essentially element-wise multiplication of $p$ permuted vectors. Such an operation heightens the difficulty against inversion at the expense of introducing distortion in the product code due to the noises involved in one or multiple permuted vectors. Thus, it is expected that the performance drops with larger $p$. This also demonstrates the common trade-off suffered in the cancelable biometrics, namely performance-security trade-off.

Apart from the above, we also examine the relation between the number of independent hashing functions $m$ and EER. Evaluation has been carried out by increasing the $m$ from 5, 10, 50, 100, 300, 500, 800 and 1000 while fixing $k$=250, and $p$=2. As expected, a better EER can be attained with respect to the increment of $m$ and level off at large $m$ as illustrated in Fig. 4(a). Since randomization takes part in the hashing functions, the randomly permuted vectors constructed is approximately Gaussian distributed with sufficiently large $m$. Experimental evidence suggests that with $m$=200 and $k$=250, we obtain reasonably good performance accuracy.

*2) Effect of the number of Gaussian random matrices m, number of Gaussian random projection vectors q*

For GRP-based IoM, we investigate the effect of the number of Gaussian random matrices $m$ and number of Gaussian random projection vectors $q$ with respect to the EER. In this experiment, $m$ is changed from 2, 5, 10, 50, 100, 150, 200, 250 and 300 while $q$ is varied from 2, 3, 5, 10, 50, 100, 150, 200, 250 and 299. Then, Fig. 5 shows the curves of "EER (%)-vs-$q$" (curves display for $m$ = 150, 200, 250 and 300 only). We can observe:

1. The $q$ plays an insignificant effect with respect to the EER. For instance, we earn EER=0.26% and EER=0.24% for $q = 2$ and 250, respectively where $m$ is fixed to 300. This observation suggests that, with an adequate $m$, $q$ can be set small without compromising EER and can greatly save computational as well as storage cost.
2. On the contrary, the number of Gaussian random matrices $m$ is a significant factor to determine accuracy performance. The experiment in FVC2002 DB1 reveals that EERs are decreased dramatically from 27.82% to 0.24% when $m$ increases from 2 to 300 for $q$ =250. This observation remains true for other datasets as shown in Fig. 4(b). This confirms the theory of LSH, where points nearby are more likely to fall in the same bucket than the points farther away after hashing. The buckets (quantization) notion in LSH indeed resembles the GRP-based IoM hashing in which the indices of the maximum value of projected vectors can be deemed as a quantization output.

*B. Accuracy Performance Evaluation*

In this section, the experiments of accuracy performance in stolen-token scenario using the best parameters tuned in the previous section are carried out for FVC2002 and FVC2004. Table II tabulates the accuracy results as well as comparisons with the baseline system and state of the arts. It can be observed that:

1. The accuracy performance of the GRP-based IoM hashing is well preserved with respect to its original fingerprint vector counterpart [29] and MCC [12] regardless in stolen-token and genuine-token cases. Furthermore, it can be observed that the accuracy difference between stolen-token and genuine-token cases is insignificant. This suggests a great advantage that the external token may not be a *secret* anymore; thus, the security and privacy attention on the *secret* requirement of the external token can be relaxed. The security and privacy analysis is provided in section VII.
2. The GRP-based IoM hashing excels the existing minutiae based fingerprint cancelable templates [13] [15] [41] in

TABLE II
PERFORMANCE ACCURACY (AVERAGE EER %) AND COMPARISON

| Methods | EER (%) for FVC2002 | | | EER (%) for FVC2004 | | |
|---|---|---|---|---|---|---|
| | DB1 | DB2 | DB3 | DB1 | DB2 | DB3 |
| **Without Template Protection** | | | | | | |
| MCC [12] | 0.60 | 0.59 | 3.91 | 3.97 | 5.22 | 3.82 |
| Fixed-length vector [29] | 0.20 | 0.19 | 2.30 | 4.70 | 3.13 | 2.80 |
| **With Template Protection** | | | | | | |
| **URP-based IoM hashing** | 0.43 | 2.10 | 6.60 | 4.51 | 8.02 | 8.46 |
| **URP-based IoM hashing - genuine token** | 0.20 | 0.88 | 1.94 | 0.44 | 3.08 | 2.91 |
| **GRP-based IoM hashing** | 0.22 | 0.47 | 3.07 | 4.74 | 4.10 | 3.99 |
| **GRP-based IoM hashing - genuine token** | 0.16 | 0.45 | 2.51 | 1.15 | 2.36 | 2.70 |
| 2P-MCC$_{64,64}$ [13] | 3.3 | 1.8 | 7.8 | 6.3 | - | - |
| Bloom Filter with fingerprint [41] | 2.3 | 1.8 | 6.6 | 13.4 | 8.1 | 9.7 |
| **Random Projection-based Approach (Same token)** | | | | | | |
| Teoh et al. [16] | 15 | 15 | 27 | - | - | - |
| Yang & Busch [7] | - | 2.65 | - | - | - | - |
| Wang & Hu [15] | 4 | 3 | 8.5 | - | - | - |

TABLE III
COMPLEXITY TO INVERT SINGLE AND ENTIRE FEATURE COMPONENT.

| Databases | Min value with four decimal precision | Max value with four decimal precision | Possibilities for single feature component | Total possibilities for entire feature |
|---|---|---|---|---|
| FVC2002 DB1 | -0.2504 | 0.2132 | $4636 \approx 2^{12}$ | $2^{12 \times 299} = 2^{3588}$ |
| FVC2002 DB2 | -0.2409 | 0.2484 | $4893 \approx 2^{12}$ | $2^{12 \times 299} = 2^{3588}$ |
| FVC2002 DB3 | -0.1919 | 0.2372 | $4291 \approx 2^{12}$ | $2^{12 \times 299} = 2^{3588}$ |
| FVC2004 DB1 | -0.2487 | 0.1748 | $4235 \approx 2^{12}$ | $2^{12 \times 299} = 2^{3588}$ |
| FVC2004 DB2 | -0.2357 | 0.1950 | $4307 \approx 2^{12}$ | $2^{12 \times 299} = 2^{3588}$ |
| FVC2004 DB3 | -0.1947 | 0.1796 | $3742 \approx 2^{11}$ | $2^{12 \times 299} = 2^{3289}$ |

terms of accuracy; this is attributed to the superior MCC descriptor and the nice property of performance preservation of fingerprint vector counterpart [29] and IoM hashing.

3. The accuracy performance of the URP-based IoM hashing decreases gradually in FVC2002 (DB1 and DB2) and FVC2004 DB1 while remains approximately 3%-5% of deterioration in the rest of the testing sets. Such deterioration is expected as the discriminate features are likely to be permuted out of the $k$-window. However, the user-specific tokens of each user counteract the loss, thus, the accuracy in genuine-token scenario is comparable to its original vector counterpart [29] and MCC [12]. This suggests that the URP-based IoM hashing demands higher discriminative features in order to preserve accuracy. Nevertheless, the accuracy of URP-based IoM hashing is still comparable to the state-of-the-arts [13] [15] [41].

### C. Discussion of the time complexity & simple realization

To investigate the time efficiency of the IoM hashing, we record both average code generation and matching time on a machine with specifications, such as Intel CPU i7-6700HQ (2.6GHz) and 8GB RAM. The timing readings are tabulated in Table IV. Note that MATLAB Ver. 2013b is used and code is not optimized. The computation time of the hashed code generation is longer than that of the matching due to a large number of independent hash functions (e.g. $m$=300, 600) are to be generated while matching takes much less time with the simple matcher described in Section V($E$). Another observation is that the generation of GRP-based hashed code is much more efficient than that of the URP-based hashed code due to the larger number of permutation. Overall, the average matching time meets the expectation for an efficient matching requirement compared to light category[1] in FVC2002 and FVC2004. We also remark that the implementation of IoM hashing is rather simple attributed to i) only random projection or random permutation is involved; ii) simple matching by mere element-wise subtraction and counting.

## VII. SECURITY AND PRIVACY ANALYSIS

We devote this section for analysis beyond accuracy, namely (1) privacy analysis, (2) security analysis and (3) revocability/unlinkability analysis. Privacy analysis refers to the feasibility of a cancelable biometric scheme to withstand any attack to recover the original biometric data. Unlike privacy attack, security attack is meant to gain the illegitimate access, with feasible attack complexity to the biometric systems by means of the fake features (pre-image) or the elaborated instances that close to the genuine biometric data.

As far as cancelable biometrics is concerned, privacy analysis includes non-invertibility and Attacks via Record Multiplicity (ARM) analysis while security analysis covers brute force analysis, ARM, false-accept analysis and birthday attack analysis.

### A. Privacy Analysis

#### 1) Non-invertibility Analysis

Non-invertibility (irreversibility) refers to the computational hardness in restoring the fingerprint vector from the IoM hashed code with and/or without permutation seeds or random matrices that correspond to URP or GRP realizations, respectively. Here, we assume the adversary manages to retrieve the hashed codes, tokens and he knows well hashing algorithm as well as the corresponding parameters (e.g. $m$, $k$, $p$, $q$). For both URP-based and GRP-based IoM that expressed in the discrete indices form, there is no clue for an adversary to guess the fingerprint vector **x** information (real-value features) directly from the *stolen hashed code alone*. Besides, knowing the token (e.g. permutation seeds or projection matrices) is also helpless in recovering the fingerprint vector as no direct link exists between token and fingerprint vector. The way for the adversary to attack is only to guess the real-value directly.

In the worst case, assume that the adversary learns the minimum and maximum values of the feature components of **x**. Let's take FVC2002 DB1 as an example, the minimum and maximum values of the feature components are -0.2504 and 0.2132 respectively. Presume the adversary attempts to guess from -0.2504, -0.2503, -0.2502 and so on, until the maximum 0.2132. There is a total of 4636 possibilities. In our implementation, the precision is fixed at four decimal digits, the possibility of guessing a single feature component of **x** requires 4636 attempts ($\approx 2^{12}$). Thus, the entire 299 feature components require around $2^{12 \times 299} = 2^{3588}$ attempts in total. The possibilities to correctly guess *a single* and *entire* feature component are presented in Table III. Obviously, this is computational infeasible.

TABLE IV
AVERAGE TIME PROCESSED IN ENROLLMENT AND MATCHING STAGES

| IoM hashing | Average Time (FVC2002) | | | Average Time (FVC2004) | | |
|---|---|---|---|---|---|---|
| | DB1 | DB2 | DB3 | DB1 | DB2 | DB3 |
| **Enrollment Stage (generating hashed code, sec)** | | | | | | |
| GRP-based IoM ($q$=16,$m$=300) | 0.0072 | 0.0075 | 0.0072 | 0.0072 | 0.0074 | 0.0070 |
| URP-based IoM ($k$=250,$m$=600) | 0.0699 | 0.0884 | 0.0781 | 0.0829 | 0.0832 | 0.0793 |
| **Matching Stage (matching hashed code, $\times 10^{-5}$ sec)** | | | | | | |
| GRP-based IoM ($q$=16,$m$=300) | 1.7238 | 1.6334 | 2.5238 | 2.2433 | 2.5394 | 1.6758 |
| URP-based IoM ($k$=250,$m$=600) | 1.8566 | 1.9421 | 2.1264 | 2.1549 | 1.9987 | 2.5874 |

---

[1] The light category is intended for algorithms conceived for light architectures and therefore characterized by low computation cost, in limited memory storage and small template size.

*2) Attacks via Record Multiplicity (ARM) for Privacy*

Attacks via record multiplicity (ARM) is a more dreadful instance of privacy attack, which utilizes *multiple* compromised protected templates with or without knowledge and parameters that associated to the algorithm for original biometric reconstruction [42]. For IoM hashing, ARM is computationally hard to infer the numerical value as the stored templates have been transformed into the rank space that is uncorrelated to the fingerprint feature space. The attack complexity is thus identical to the non-invertibility attack presented in Table III.

B. *Security Attacks Analysis*

*1) Brute-Force Attack*

Brute-force attack is an instance of security attacks. It is also known as pre-image attack or masquerade attack [5] in the literature. For URP realization, we take *m*=600 and *k*=128, which is the best accuracy configuration observed from section VI(*A*). Since the indices of IoM hashed code taking a value between 1 and 128. Hence, the guess complexity for each entry is $k = 128 = 2^7$ and thus 600 entries requires $2^{4200}$ attempts, which are far beyond to be reached by the present computing facility. On the other hand, for GRP realization with *m*=300 and *q*= 64, the attack complexity for each entry is $q = 150 > 2^7$, the complexity for 300 entries requires $2^{2100}$ attempts that is again computational infeasible.

Furthermore, we verify the above theoretical estimation empirically in sequel. In the experiment, we randomly generate a large number of hashed codes and perform illegitimate access. For each fingerprint vector, 1000 randomly hashed codes are generated and match against the fingerprint vector with identical permutation seeds or projection matrices. This results 500,000 (5×100×1000) matching scores in which we named *brute-force attack scores*. We plot the imposter and brute-force attack distribution as shown in Fig. 6. It can be observed that the brute-force attack distribution is either strongly overlapped or offset to the left with respect to the imposter distribution. This indicates that the brute-force attack scores lead to an equivalent dissimilarity with the imposter scores. This experimental result echoes the theoretical estimation above; thus, conclude the infeasibility of the brute-force attack.

*2) Attacks via Record Multiplicity (ARM) for Security*

Unlike ARM for privacy, ARM is also a plausible security attack where *multiple* compromised protected templates with or without knowledge on the parameters that associated to the algorithm are utilized to generate a pre-image instance. In our case, if knowing the order of the feature components (not necessary the numerical value) and the permutation seeds, an elaborate faked feature vector can be formed. Consequently, the largest value resulted from the product of two permuted feature vectors appears in the desired position. The system can thus be broken.

Now, the attack complexity of ARM can be perceived as the complexity of determining the order of the feature components in the fingerprint vector. For instance, for URP-based realization, let $\mathbf{x} = \{x_a, x_b, x_c\}$ be the input features, with $p = 2$ and two randomly permuted feature vectors $\hat{\mathbf{x}}_1 = \{x_c, x_a, x_b\}$ and $\hat{\mathbf{x}}_2 = \{x_b, x_c, x_a\}$. The Hadamard product vector $\bar{\mathbf{x}} = \{x_b x_c, x_a x_c, x_a x_b\}$ can be computed. Assume $x_a x_c$ is the largest component, i.e. $x_a x_c = \max(\prod_{l=1}^{2}(\hat{\mathbf{x}}_l(j)))$ $j = 1,2,3$, two inequalities can be derived as follows: $x_a x_c > x_b x_c$ and $x_a x_c > x_a x_b$. We can further reason that for $x_a > x_b$ and $x_c > x_b$, the adversary can retrieve the entire order information (e.g. $x_a > x_c > x_b$) by repeating this process with multiple hashed codes. However, in practice, our feature components contain both positive and negative values. For instance, let $\mathbf{x} = \{x_a, x_b, x_c\} = \{$-0.2, 0.5, -0.1$\}$, by using $x_a x_c > x_b x_c$, ie. $(-0.2) \times (-0.1) > (0.5) \times (-0.1)$, we obtain $(-0.2) < (0.5)$, hence $x_a < x_b$. This is contradicted with $x_a > x_b$.

Such inequality reasoning is applicable to GRP-based realization as well attributed to the multiplication is a common ingredient for both Hadamard product in URP-based realization and the inner product (random projection) in GRP-based realization. Therefore, ARM for IoM hashing is only valid if biometric feature values are either all negative or positive, the mixed sign in feature values renders inequality reasoning indefinite. We can conclude that the inequality relation is insufficient to launch ARM as a means of security attack.

*3) False Accept Attack*

Unlike blind guessing on the entire hashed code in brute force attack, false accept attack (dictionary attack) may requires far less number of attempts to gain illegitimate access [43]. In fact, this attack is viable due to the fact that thresholding-based decision scheme is commonly applied in biometric systems. In other words, access would be granted as long as the matching score succeeds the pre-defined threshold $\tau$. Let we take the best performing parameters from the FVC2002DB1 experiments in section VI(*A*), i.e. *m*=600 and *k*=128, the decision threshold $\tau$ when FAR =FRR is observed is 0.11. Hence, the minimum number of matched (collided) entries in IoM hashed codes pair for successful access is $\tau \times m = 0.11 \times 600 = 66$. The window size *k* indicates 128 possible indices values that is equivalent to $2^7$ $(= 2^{\log_2 k})$ guessing effort is required for an entry. Therefore, the false

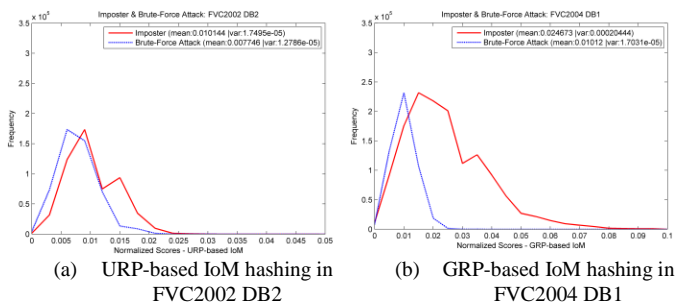

(a) URP-based IoM hashing in FVC2002 DB2  (b) GRP-based IoM hashing in FVC2004 DB1

Fig. 6 Imposter and brute-force attack distribution for brute-force attack analysis

TABLE V
FALSE ACCEPT ATTACK COMPLEXITY ANALYSIS FOR IOM HASHING.

| Databases | URP-based IoM hashing | | | | | GRP-based IoM hashing | | | | |
|---|---|---|---|---|---|---|---|---|---|---|
| | $\tau$ | $m$ | $\tau \times m$ | $k$ | Minimum attack complexity | $\tau$ | $m$ | $\tau \times m$ | $q$ | Minimum attack complexity |
| FVC2002 DB1 | 0.11 | 600 | 66 | $128 = 2^7$ | $(2^7)^{66} = 2^{462}$ | 0.01(_0.06_) | 300 | 3 (_18_) | $16 = 2^4$ | $2^{12}$ ($2^{72}$) |
| FVC2002 DB2 | 0.08 | 600 | 48 | $250 \approx 2^8$ | $(2^8)^{48} = 2^{384}$ | 0.01(_0.06_) | 300 | 3 (_18_) | $16 = 2^4$ | $2^{12}$ ($2^{72}$) |
| FVC2002 DB3 | 0.05 | 600 | 30 | $250 \approx 2^8$ | $(2^8)^{30} = 2^{240}$ | 0.01(_0.05_) | 300 | 3 (_15_) | $16 = 2^4$ | $2^{12}$ ($2^{60}$) |
| FVC2004 DB1 | 0.10 | 600 | 60 | $50 > 2^5$ | $(2^5)^{60} = 2^{300}$ | 0.01 | 300 (_1200_) | 3 (_12_) | $150 > 2^7$ | $2^{21}$ ($2^{91}$) |
| FVC2004 DB2 | 0.06 | 600 | 36 | $250 \approx 2^8$ | $(2^8)^{36} = 2^{288}$ | 0.01 | 300 (_1200_) | 3 (_12_) | $150 > 2^7$ | $2^{21}$ ($2^{91}$) |
| FVC2004 DB3 | 0.06 | 400 | 24 | $250 \approx 2^8$ | $(2^8)^{24} = 2^{192}$ | 0.01 | 300 (_1200_) | 3 (_12_) | $150 > 2^7$ | $2^{21}$ ($2^{91}$) |

accept attack complexity can be estimated from $(2^{\log_2 k})^{\tau \times m}$.

For URP-based IoM hashing, we observed from TABLE V that the attack complexity reduced to $2^{462}$ compare to $2^{4200}$ in brute force attack for FVC2002DB1 and the complexity reduction appears identical for the rest of testing data sets. However, we note that the reduced attack complexity is still favorably high to resist false accept attack.

On the other hand, the attack complexity of GRP-based IoM hashing is more critical with carelessly selected parameters as evidenced in TABLE V. To escalate the attack difficulty, we can increase either $m$ or $\tau$. Let $m = 1200$ be set for FVC2004 (DB1, DB2 and DB3), the attack complexity thus, reaches a level of computational infeasibility. Another option is to elevate $\tau$ by reducing the $q$ as shown in TABLE V (FVC2002 DB1, DB2 and DB3). This option is preferable than the former i.e. increasing $m$ as it reduces the number of random Gaussian projection vectors. On the contrary, the former option requires more storage space for large $m$.

*4) Birthday Attack*

Apart from the above attacks, we present a new security attack for cancelable biometrics, namely *Birthday attack* [44]. Specifically, a birthday attack exploits the mathematics behind the birthday problem [44] in probability theory. The attack relies on the higher collision likelihood found between the random attacks and a fixed degree of permutations (pigeonholes).

In our context, birthday attack refers to a scenario where an adversary has gained a large number of hashed codes from the compromised databases. This leads to a plausible security attack where at least a collision (match) can be found between any two hashed codes from $N_t$ hashed codes where $N_t \gg 1$. For the hashed code with a single entry ($m = 1$), the *expected* trials of finding the *first* collision is $Q(\delta) = \sqrt{\frac{\pi}{2}\delta}$, where $\delta$ is the largest entry value of IoM hashed codes, i.e. $\delta = k$ for URP and $\delta = q$ for GRP. Suppose there is a collision of two hashed codes i.e. $h_i(X) = h_i(Y)$ where $i = 1, \ldots, m$ and $h(X), h(Y) \in [1, \delta]$. The *expected* trials of finding the collisions for $\tau m$ elements is $\left(\frac{\delta\pi}{2}\right)^{\frac{\tau m}{2}}$.

From TABLE VI, it can be observed that the complexity of birthday attack over the conventional false-accept attack is reduced approximately in square root degree. More importantly, the TABLE VI reminds us that attack difficulty is closely associated to the parameters selected. For instance, the complexity on FVC2004 DB3 is $2^{35}$ that are not ideally secure. Yet, we can further increase the complexity by either increasing $m$ or reducing $\tau$ without compromising the performance accuracy. Concisely, a considerate justification is required to ensure the desired security level as well as performance criteria with proper parameters tuning.

*C. Non-linkability*

In this section, the requirement of non-linkability (unlinkability) is validated. To do this, we introduce the *pseudo-genuine scores*. The pseudo-genuine score refers to the matching scores between the hashed codes generated from different fingerprint templates of the same individual by using different permutation seeds or projection matrices. This protocol resembles the genuine matching protocol that yields 1000 pseudo-genuine matching scores as well. Refer to section VII(*D*), the pseudo-imposter score is computed between two hashed codes generated from each template using different permutation seeds or projection matrices. In this context, when the pseudo-imposter and pseudo-genuine distribution are overlapped, it implies that the hashed codes generated from the same user or from the others are sufficiently indistinctive. On the contrary, if both distributions are far separated, this allows the adversary to differentiate the hashed code easily whether it is generated from the identical individual. The difficulty in differentiating the hashed codes contributed to the non-linkability or unlinkability property. Fig. 7 illustrates the pseudo-imposter and pseudo-genuine distribution plot where the pseudo-imposter and pseudo-genuine distribution are largely overlapped. This supports that the IoM hashed codes satisfy non-linkability or unlinkability property.

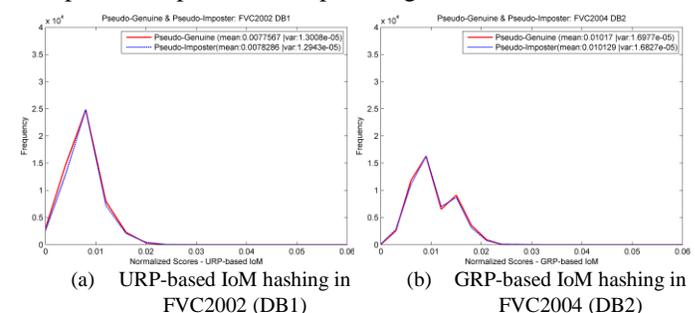

| (a) URP-based IoM hashing in FVC2002 (DB1) | (b) GRP-based IoM hashing in FVC2004 (DB2) |

Fig. 7 Pseudo-Genuine & Pseudo-Imposter distributions for non-linkability analysis. Overlapped distributions indicates indistinctiom of hashed codes that generated from the same user or from the others.

TABLE VI
BIRTHDAY ATTACK COMPLEXITY ANALYSIS FOR IOM HASHING.

| Databases | URP-based IoM hashing | | | | | GRP-based IoM hashing | | | | |
|---|---|---|---|---|---|---|---|---|---|---|
| | $\tau$ | $m$ | $\tau \times m$ | $k$ | *expected* trials: $\left(\frac{k\pi}{2}\right)^{\frac{\tau m}{2}}$ | $\tau$ | $m$ | $\tau \times m$ | $q$ | *expected* trials: $\left(\frac{q\pi}{2}\right)^{\frac{\tau m}{2}}$ |
| FVC2002 DB1 | 0.11 | 600 | 66 | 128 | $2^{252}$ | 0.06 | | 66 | | $2^{42}$ |
| FVC2002 DB2 | 0.08 | 600 | 48 | 250 | $2^{207}$ | 0.06 | | 48 | | $2^{42}$ |
| FVC2002 DB3 | 0.05 | 600 | 30 | 250 | $2^{130}$ | 0.05 | 300 | 30 | 16 | $2^{35}$ |
| FVC2004 DB1 | 0.10 | 600 | 60 | 50 | $2^{189}$ | 0.10 | | 60 | | $2^{70}$ |
| FVC2004 DB2 | 0.06 | 600 | 36 | 250 | $2^{156}$ | 0.06 | | 36 | | $2^{42}$ |
| FVC2004 DB3 | 0.06 | 400 | 24 | 250 | $2^{104}$ | 0.05 | | 24 | | $2^{35}$ |

*D. Revocability*

In this section, revocability (renewability) is evaluated by conducting the experiments described in [45]. This produces $100 \times (5 \times 100) = 50000$ *pseudo-imposter* scores. The genuine, imposter and pseudo-imposter distributions are shown in Fig. 8 where FVC2002 is used for URP-based IoM hashing while FVC2004 is tested for GRP-based IoM. From Fig. 8, it is observed that a large degree of overlapping occurs between the imposter and pseudo-imposter distributions. This implies the newly generated hashed codes with the given 100 random permutation sets and/or random projection matrices are distinctive even though it is generated from the identical source of fingerprint vector. In terms of verification performance, we obtain EER = 0.23%, 0.98%, 1.77% for URP-based IoM hashing in FVC2002 (DB1, DB2, DB3) and 1.38%, 2.32%, 2.56% for GRP-based IoM hashing in FVC2004 (DB1, DB2, DB3) respectively in which intersection of genuine and pseudo-imposter distribution is taken. This verifies that the proposed both IoM realizations satisfies the revocability property requirement. We further note that the experiment presented in section VI(*B*) suggests that the stolen tokens (e.g. permutation seed or random matrices) would not compromise the accuracy performance significantly. Thus, tokens in IoM hashing merely serves for the revocability and are not secret for public.

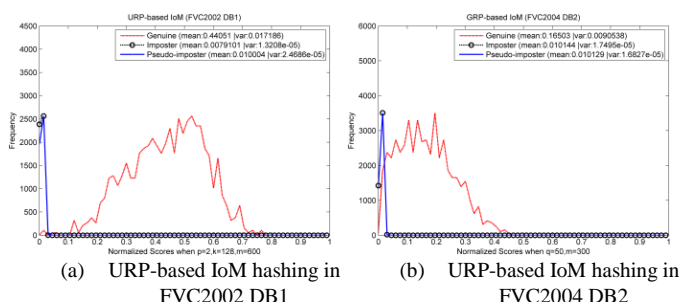

(a) URP-based IoM hashing in FVC2002 DB1  (b) URP-based IoM hashing in FVC2004 DB2

Fig. 8 The genuine, imposter, and pseudo-imposter distributions for revocability analysis. A large degree of overlapping of pseudo-imposter and imposter distributions illustrates that hashed codes generated using different random vectors leads to significant distinction.

VIII. CONCLUSION AND FUTURE PROSPECTS

In this paper, we present a generic ranking based LSH motivated two-factor cancelable biometrics scheme dubbed IoM hashing. We demonstrate two realizations of the IoM hashing for fingerprint biometrics, namely URP-based and GRP-based IoM hashing. From both theoretical justification and empirical observations, the IoM hashed codes can largely preserve accuracy performance with respect to its before-transformed counterparts thanks to the nice property of localized random projection and the rank correlation for GRP hashing and URP hashing, respectively. The IoM hashing is also shown satisfy both non-linkability and revocability template protection criteria. The IoM hashing is strongly resilient against the existing several major security and privacy attacks subject to the properly tuned parameters. Finally, we outline a new attack for cancelable biometrics namely birthday attack, which reveals even more dreadful than the dictionary attack (false accept attack). Our future work consists of four directions to further push the limits of the IoM hashing. The first direction is to extend the work to unordered variable-sized representation such as fingerprint minutiae. The second direction is to integrate with biometric cryptosystem primitives, e.g. Fuzzy Vault, Fuzzy Commitment, where secret-key vs privacy-leakage framework [46] could be incorporated for privacy preserving analysis. That would be a strong complement for cryptographic keys generation and protection purposes. Lastly, as a two-factor cancelable biometric scheme, the IoM hashing is naturally fitted to identity authentication scenario. However, it would be interesting to investigate the possibility of using IoM hashing in identification setting.

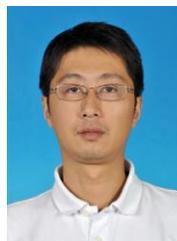

**Zhe Jin** obtained his BIT (Hons) in Software Engineering, MSc (I.T.) from Multimedia University, Malaysia in 2007 and 2011 respectively, and Ph.D. in Engineering from University Tunku Abdul Rahman Malaysia in 2016. Currently, he is a Lecturer at School of Information Technology, MONASH University, Malaysia campus. His research interest is Biometric Template Security.

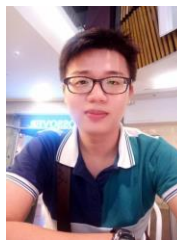

**Yen-Lung Lai** obtained his BSc degree in Physics from University Tunku Abdul Rahman (UTAR), Malaysia in 2015. Currently, he is pursuing a Ph.D. at Monash University Malaysia. His research interests include Biometrics security, and Cryptography.

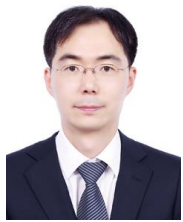

**Jung Yeon Hwang** received the BS in mathematics from Korea University, the MS and Ph.D. in information security from Korea University, Seoul, in 1999, 2003, and 2006, respectively. He was a visiting researcher at Columbia University (in the city of New York) from Oct. 2005 to April 2006 and post-doctoral researcher at Korea University from 2006 to 2009. Since 2009, he has been a senior researcher of Electronics and Telecommunications Research Institute (ETRI), Daejeon, Korea. He has developed broadcast encryption schemes and anonymous signature schemes with both opening and linking capability. His research interests include biometrics, fuzzy cryptosystem, identity management, privacy-enhancing technology, cryptographic protocols, and their applications.

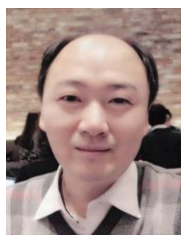

**Soohyung Kim** received the B.S. and M.S. degrees in computer science from Yonsei University, Seoul, Korea, in 1996 and 1998. He received the Ph.D. degree in computer science from Korea Advanced Institute of Science and Technology (KAIST), Daejeon, Korea in 2016. He is currently a project leader of Information Security Research Division in Electronics and Telecommunications Research Institute (ETRI), Daejeon, Korea. His research interests include biometrics, identity management, payment system, network and system security.

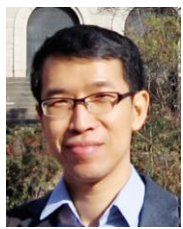

**Andrew Beng Jin Teoh** (SM'12) obtained his BEng (Electronic) in 1999 and Ph.D. degree in 2003 from National University of Malaysia. He is currently an associate professor in Electrical and Electronic Engineering Department, College Engineering of Yonsei University, South Korea. His research, for which he has received funding, focuses on biometric applications and biometric security. His current research interests are Machine Learning and Information Security. He has published more than 270 international refereed journal papers, conference articles, edited several book chapters and edited book volumes. He served and is serving as a guest editor of IEEE Signal Processing Magazine, associate editor of IEEE Biometrics Compendium and editor-in- chief of IEEE Biometrics Council Newsletter.